\documentclass[final]{cvpr/cvpr}

\usepackage[title]{appendix}

\usepackage{times}
\usepackage{epsfig}
\usepackage{graphicx}
\usepackage{amsmath}
\usepackage{amssymb}
\usepackage{booktabs}
\usepackage{xcolor}
\usepackage{multirow}

\usepackage[breaklinks=true,bookmarks=false]{hyperref}

\begin{document}

\title{Learning Video Representations from Textual Web Supervision}

\author{Jonathan C. Stroud$^{1, 2}$\thanks{Contact: \texttt{stroud@google.com}} \and Zhichao Lu$^{2}$ \and Chen Sun$^{2}$ \and Jia Deng$^{3}$ \and Rahul Sukthankar$^{2}$ \and Cordelia Schmid$^{2}$ \and David A. Ross$^{2}$ \and \\ $^{1}$University of Michigan \hspace{5mm} $^{2}$Google Research \hspace{5mm} $^{3}$Princeton University}

\maketitle

\begin{abstract}
Videos on the Internet are paired with pieces of text, such as titles and descriptions. This text typically describes the most important content in the video, such as the objects in the scene and the actions being performed. Based on this observation, we propose to use text as a method for learning video representations. To accomplish this, we propose a data collection process and use it to collect 70M video clips shared publicly on the Internet, and we then train a model to pair each video with its associated text. We evaluate the model on several down-stream action recognition tasks, including Kinetics, HMDB-51, and UCF-101. We find that this approach is an effective method of pre-training video representations. Specifically, it outperforms all existing methods for self-supervised and cross-modal video representation learning.
\end{abstract}

\section{Introduction}

Video representations are typically learned in a fully-supervised fashion. For this approach to be successful, we require large amounts of labeled data, typically on the order of hundreds of thousands of labels. Acquiring these labels can cost tens of thousands of hours of human time to annotate~\cite{gu2018ava}, and furthermore, when datasets become large, the benefit of gathering more labels appears to diminish~\cite{allegro2019}. At a certain point, it becomes too costly to simply label more data to improve performance. In this regime, we look to alternative sources of supervision to learn video representations without costly manual labels.

In our work, we draw this supervision from textual metadata available publicly on the Internet. Specifically, we use videos from YouTube, where videos are associated with freeform text in the form of titles, descriptions, tags, and channel/creator names. These four pieces of textual metadata provide rich information about each video's content. Frequently, they describe the exact types of information which labelers are asked to annotate in labeled datasets, such as objects, scenes, and human actions. For example, consider the title, ``Learning how to swim!'' or the channel name ``PotteryMaker''. Both of these indicate what actions will take place in their respective videos, and we can leverage this information to learn representations, in much of the same way we use labels in supervised learning.

One advantage of using text from internet videos is that the text is open-ended and therefore more descriptive than class labels. Consider the title ``Outdoor free-climbing in Yosemite''. In supervised learning, this video would be labeled ``rock climbing'', but this label ignores important information about the scene and the specific type of action, potentially missing out on valuable supervisory signal. In our experiments, we demonstrate that using text, and using multiple sources of text, translates into impressive downstream performance, without the need for class labels during pre-training. We compare our method with other self-supervised and cross-modal approaches, showing that our method produces video representations which improve downstream performance by 8.9\% on HMDB-51~\cite{kuehne2011hmdb} and 2.1\% on UCF~101~\cite{soomro2012dataset} (Section~\ref{sec:experiments}).

Another advantage of this approach is that the amount of available data is immense; e.g.\ over 500 hours of content is uploaded every minute to YouTube alone~\cite{tubefilter2019}, and each video is labeled with text. To leverage this data, we propose a data collection process called Weak Textual Supervision (WTS). We use a text-based video search engine to query for common words and collect a large-scale video dataset with matched pieces of text metadata. Using this process, we collect a dataset of 70 million videos, which is comparable in scale to the recent HowTo100M dataset~\cite{miech2019howto100m}, but includes paired metadata, is drawn from a higher number of unique videos (70M vs 1M), and is not limited to only instructional videos (Section~\ref{sec:dataset}). Few truly large-scale video datasets like ours are currently available (Table~\ref{tab:datasets}), and even fewer are available publicly. We intend to release the list of videos used in our experiments to facilitate future research.

Our goal with this data is to learn video representations, that is, feature vectors which encode a video clip, which are then useful for downstream tasks. To learn these representations, we propose a training scheme in which the video representation are used to pair each video with its associated metadata. We use powerful 3D Convolutional Neural Network (3D CNN) architectures to produce these representations, and we train the video representations end-to-end on WTS-70M (Section~\ref{sec:method}). We evaluate the representations' effectiveness on a suite of downstream tasks. We find that, for both fine-tuned and frozen embeddings, pre-training with our approach significantly improves downstream performance, achieving state of the art results. This is particularly true in low-data regimes, but we also show that our pre-training is useful in settings where is complementary to strongly-supervised pre-training (Section~\ref{sec:experiments}).

Our key finding is that textual metadata is a rich source of supervision which can be acquired freely from public sources. Specifically, we make the following contributions:

\begin{itemize}
\item We propose a data collection process (WTS), which uses text-based search to gather a large-scale dataset of video clips and their associated metadata, including titles, descriptions, tags, and channel names.
\item We release a dataset of 70M videos collected with this process, including their associated metadata, and pre-trained models used in this work.
\item We propose a method for learning video representations by learning to match these representations with their associated metadata, and demonstrate that our approach outperforms all other pre-training approaches.
\end{itemize}

\section{Related Work}

\noindent \textbf{Unsupervised and Self-Supervised Learning.} Many prior works have learned video representations without manual labels. In unsupervised and self-supervised learning, the supervision instead comes purely from the video itself. For example, prior approaches have successfully leveraged supervision from clip and frame ordering~\cite{misra2016shuffle,fernando2017self,lee2017unsupervised,wei2018learning,xu2019self,kim2019self}, geometry~\cite{gan2018geometry,jing2018self}, motion~\cite{pathak2017learning,lai2019self}, colorization~\cite{vondrick2018tracking}, cycle consistency~\cite{dwibedi2019temporal,wang2019learning}, video prediction~\cite{mathieu2015deep,lotter2016deep,vondrick2016generating,vondrick2016anticipating,wang2019self}, tempo~\cite{yang2020video}, and in-video consistency~\cite{gordon2020watching,qian2020spatiotemporal}. Generally, these approaches are outperformed by those leveraging supervision from other modalities or from web-based metadata. In addition, they have the disadvantage that they typically use curated datasets for pre-training, and ignore the existing labels from the dataset. In our approach, we do not use pre-labeled videos as our source of videos. These videos provide a more realistic reflection of the videos available in the wild.

\noindent \textbf{Webly-Supervised Learning.} Many prior works have leveraged webly-labeled data for visual representation learning, both for images as well as videos. In general, these approaches use metadata found on the Internet to infer weak labels for a set of images or videos, and they differ in how these weak labels are created. The most commonly-used approach is to use image search results, and to label each image with the query that was used to find it~\cite{wang2008annotating,fergus2010learning,schroff2010harvesting,bergamo2010exploiting,chen2013neil,divvala2014learning,gong2014improving,chen2015webly,gan2016webly,kuehne2019mining}. Another approach is to use text captions, and label each image with key words present in the caption~\cite{ordonez2011im2text,joulin2016learning,li2017learning,mithun2018webly}. Other approaches use user-defined keywords or tags~\cite{gong2014multi,izadinia2015deep,mahajan2018exploring,ghadiyaram2019large} or algorithmically-generated topics~\cite{abu2016youtube,karpathy2014large} to the same end. These approaches have consistently demonstrated that webly-supervised learning is scalable and that it improves performance on downstream tasks, suggesting that webly-acquired class labels provide valuable supervision.

\noindent \textbf{Cross-Modal Learning.} Several prior works have used other modalities, such as text or audio, as a source of supervision for video. This approach is convenient because videos are almost always paired other modalities such as audio, and often works well because these other modalities are tightly correlated with what is happening in the video. Prior works have leveraged ambient sound~\cite{owens2016ambient,aytar2016soundnet,arandjelovic2017look,zhao2018sound,korbar2018cooperative,owens2018audio,rouditchenko2019self}, dialogue~\cite{nagrani_cvpr_2020}, and narration~\cite{yu2014instructional,alayrac2016unsupervised,zhou2018towards,zhukov2019cross,miech2019howto100m,miech2019end,alwassel2019self}, all of which of which serve as useful signals. Those approaches using narration typically do so with instructional videos, such as in the recent HowTo100M dataset~\cite{miech2019howto100m}, since instructional videos typically contain narration which describes the actions being performed. Another recent approach, like ours, uses text data which is paired with videos, such as titles~\cite{li2020learning}. This approach reaps the benefits offered by rich, descriptive supervision, and can be used with any genre of videos. Our work differs in that we also use other forms of metadata, such as descriptions. In addition, this prior work uses curated data from Kinetics-400, while we introduce WTS as our source of videos, which is more representative of videos in the wild.

\section{Data Collection}
\label{sec:dataset}

\begin{figure*}[t!]
\centering
\includegraphics[width=.45\textwidth]{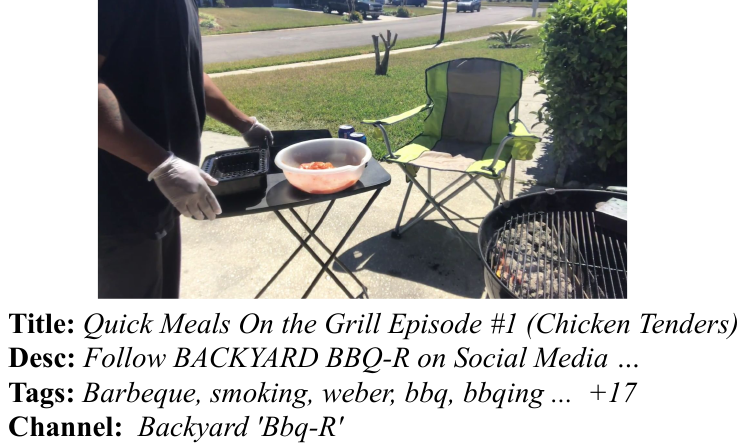} \quad
\includegraphics[width=.45\textwidth]{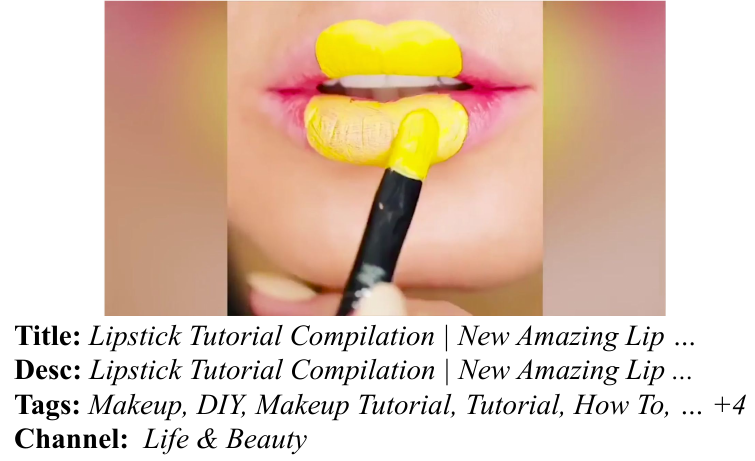}
\caption{Examples of video frames and metadata collected using WTS. Metadata typically contains references to actions (\emph{bbqing}) as well as objects (\emph{lipstick}) which are present in the scene. We collect four types of metadata for each video: titles, descriptions, tags, and channel names. Metadata is truncated where necessary for ease of visualization. All videos used under CC BY 2.0 license.}
\label{fig:datasetexamples}
\end{figure*}

We propose a data collection process in which we search for common action categories using a text-based video search engine. We begin by manually selecting the set of action categories; in our experiments we use the 700 action categories in Kinetics-700~\cite{carreira2019short}. We choose these categories primarily because they are designed to cover a broad range of human actions. In addition, this choice helps make a fair comparison with fully-supervised pre-training on Kinetics when evaluating on down-stream tasks such as HMDB-51 and UCF101, since we know that the change in performance is not just due to a different choice of action categories. Using the Kinetics classes may also help improve performance on Kinetics itself. In practice, this is a useful feature of using videos from the internet: researchers who reproduce this pipeline may opt to use a different set of action categories to better match the downstream tasks that they are targeting. Once we have these action categories, we search YouTube for these terms and collect the resulting videos. We then apply two selection criteria to filter videos. First, we discard videos which are less than 10 seconds long, since we use 10-second clips during training. Second, we discard videos which were uploaded in the past 90 days, because newer videos are more likely to be deleted, improving reproducibility of our experiments. In total, we collect 100K videos from each of the 700 queries, resulting in a dataset of 70M videos. From each video, we randomly select a 10-second clip to download.

Each video is paired with four pieces of textual metadata: its title, description, tags, and channel name. These were chosen for two reasons. First, they are all manually written by the user, as opposed to being automatically generated. Second, from manual inspection, we see that these pieces of text consistently contain informative references to content in the video. These references are written deliberately by the user, who generally will choose a title, description, and tags which help other users find their video. The user will also select a channel name (an identifier used to represent the user) which is informative, typically one which is indicative of the types of videos that the channel contains. The channel name provides context which the other signals may not, for example, a channel for guitar lessons, ``Jeff's Guitar Lessons'',  may not say ``guitar lesson'' in each video title, but the channel name makes this obvious. For some examples of videos and their metadata, see Figure~\ref{fig:datasetexamples}.

We note that WTS is a data collection process, and the datasets used in this work (denoted WTS-X) are not intended to serve as static datasets. This is made possible by the fact that our data collection process is entirely automatic, and does not rely on any manual annotation. Therefore, it is possible for WTS to be repeated or expanded flexibly with any desired action vocabulary, depending on the needs of the downstream tasks. Our experiments show that, even with 70M videos, WTS has not plateaued in terms of performance gains, suggesting that a static dataset would be limiting. Our non-static dataset provides an additional advantage over large-scale labeled datasets (such as Kinetics~\cite{kay2017kinetics}), in which videos can be deleted by their owners at any time. When a labeled video is deleted, it leads to a decay in the number of labeled videos, but in our case, no videos are labeled, so we can simply repeat WTS to account for the lost videos.

One additional advantage of using unlabeled videos is that it results in a dataset that is less curated than those which were originally labeled but had the labels stripped from them. While it is quite common to use labeled datasets in this way for unsupervised learning, these datasets are guaranteed to not have examples outside of the class label vocabulary, and it is not clear whether the results will generalize to uncurated datasets where this assumption does not hold. Our dataset does not require any labels to construct at any step of the process, and therefore it does not have this issue. Our dataset is not purely uncurated, since we intentionally use a targeted set of action classes to search for videos (as do other uncurated datasets, including HowTo100M~\cite{miech2019howto100m}). However, it is not curated in a way that requires any manual human effort.

In Table~\ref{tab:datasets}, we compare WTS-70M to other unlabeled datasets for video representation learning. In terms of the number of videos, WTS-70M is on par with the largest datasets in prior work, with 5M more unique source videos than~\cite{ghadiyaram2019large}. We acknowledge that, conceptually, any of these prior datasets could be scaled to much larger sizes simply by collecting more data, making dataset size a dubious method of comparison. However, it is still important to study how these methods behave when scaled to extreme dataset sizes, and therefore our experiments on 70M videos are a valuable contribution in this space. These experiments are particularly important because there are non-trivial issues associated with scaling unsupervised learning to extreme dataset sizes. The key issue is that we use search results to collect data, and the quality of these results declines as we move deeper into the search rankings to collect more videos. Despite this, we demonstrate that performance does not saturate by the time we hit 70M videos (Section~\ref{sec:scalingexps}).

\setlength{\tabcolsep}{4pt}
\begin{table}[t]
\begin{center}
\begin{tabular}{lccc}
\toprule
Dataset & Videos & Duration (hrs) & Supervision \\
\midrule [0.1em]
Sports-1M~\cite{karpathy2014large} & 1.1M & 15K & Topics \\
Youtube-8M~\cite{abu2016youtube}  & 8M & 500K & Topics \\
HowTo100M~\cite{miech2019howto100m}  & 1.2M & 136K & Speech \\
IG-Kinetics~\cite{ghadiyaram2019large}  & 65M & 72K & Hashtags \\
\noalign{\smallskip}
\hline
\noalign{\smallskip}
WTS-70M (ours) & 70M & 194K & Metadata \\
\bottomrule [0.1em]
\end{tabular}
\end{center}
\caption{Datasets for video representation learning. WTS-70M contains 70 million clips, each from a unique source video, and each video is paired with textual metadata.}
\label{tab:datasets}
\vspace{1mm}
\end{table}
\setlength{\tabcolsep}{1.4pt}

\begin{figure*}[t]
\centering
\includegraphics[height=.33\textwidth]{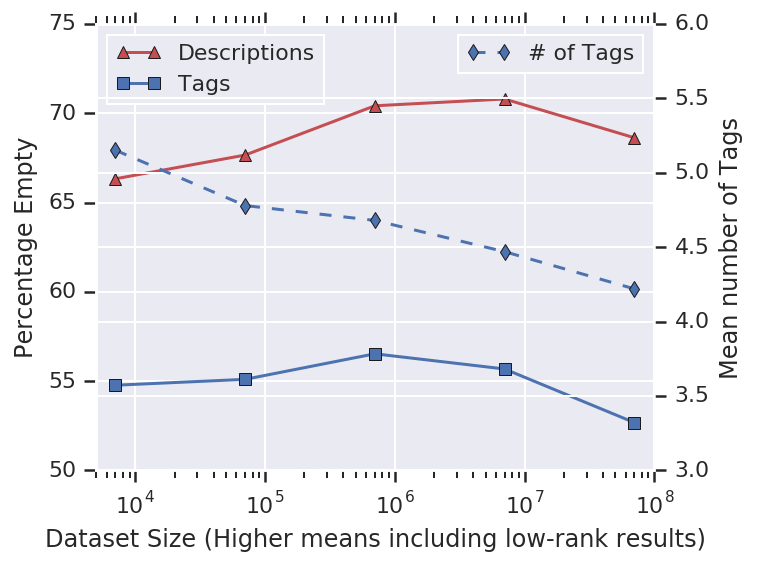} \quad
\includegraphics[height=.33\textwidth]{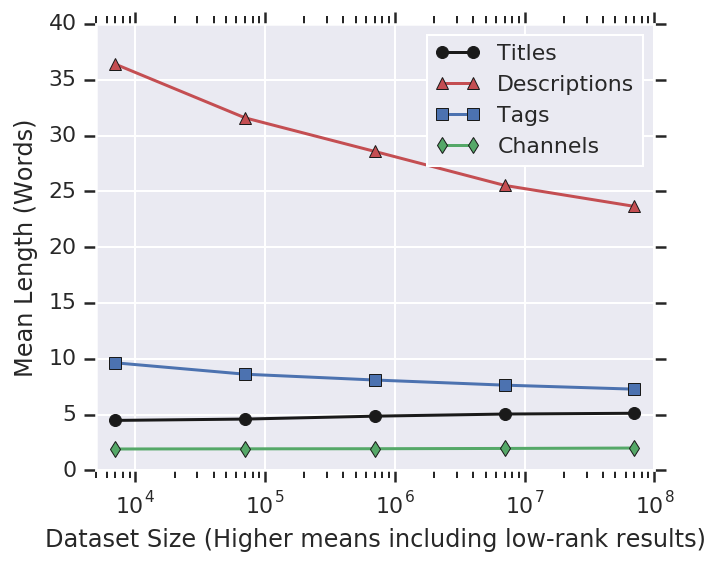}
\caption{Scaling properties of WTS. \textbf{Left:} Rate of missing descriptions and tags, and number of tags. Both descriptions and tags are empty for a large number of videos, at all dataset sizes. \textbf{Right:} Mean length (in words) of each metadata type. Descriptions and tags tend to get shorter with larger dataset sizes, but titles and channel names tend to get slightly longer.}
\label{fig:scalinganalysis}
\end{figure*}

To analyze the scaling properties of WTS, we collect increasingly-large subsets of the dataset and measure indicators of their quality, shown in Figure~\ref{fig:scalinganalysis}. The dataset size is scaled up as one would do in practice, by selecting more and more of the top search results from each query, rather than by performing a random sample from the full WTS-70M dataset. The indicators measure, for each piece of metadata, the mean length (in words), the rate of missing-ness (for descriptions and tags, which can be omitted by the user), and the mean number of tags. We find that search results are imbalanced in terms of how these indicators are distributed. Specifically, descriptions and tags tend to get \emph{shorter} with larger dataset sizes, but titles and channel names in fact get \emph{longer}. We also find that the percentage of videos which have any tags or a description stays relatively constant, but the average number of unique tags drops. These analyses indicate that the quality of descriptions and tags tend to decrease, that is, they get shorter and therefore less descriptive, for larger dataset sizes. Notably, we do not see the same for titles or channel names, indicating that these may be a more reliable source of supervision at the largest dataset sizes. This is reflected in our experiments in Section~\ref{sec:scalingexps}, where we find that using all sources of metadata is helpful for smaller dataset sizes, but that these additional sources of metadata reduce performance when scaled to the largest dataset sizes.

\noindent \textbf{Implementation Details and Deduplication.} Since Kinetics videos are also collected from the Internet, we discard videos from WTS which appear in the Kinetics validation or test sets. Since many videos do not contain a description or tags, we code the missing information as an empty string, rather than discarding these videos. We perform all searches in English, so WTS contains primarily (though not exclusively) English-language videos and metadata. However, our approach is extensible to any language.

\section{Model}
\label{sec:method}

\begin{figure}[t]
\centering
\includegraphics[width=0.45\textwidth]{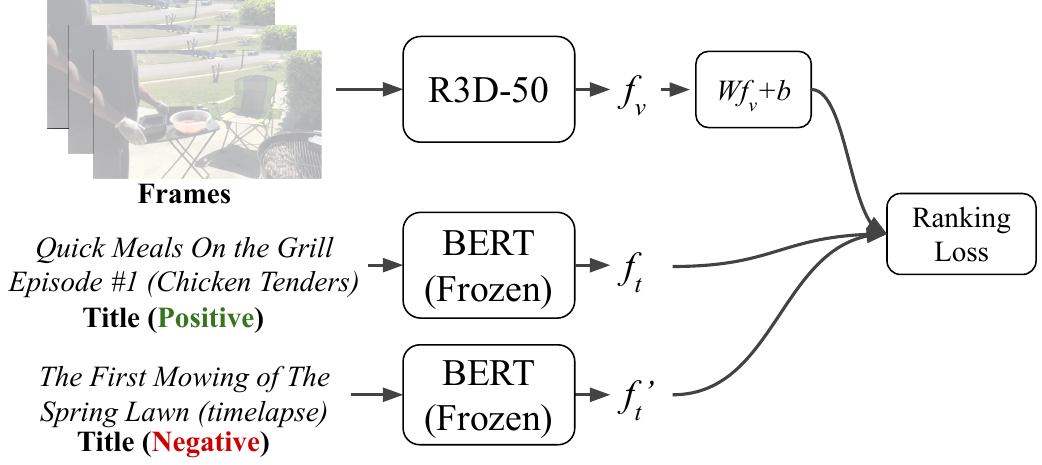}
\caption{Model architecture for cross-modal unsupervised learning from textual metadata. We encode the video using R3D-50~\cite{he2016deep}, and the metadata using BERT~\cite{devlin2018bert}. We then train the video representation by matching it with the correct metadata representation.}
\label{fig:model}
\end{figure}

At a high level, our approach (Figure~\ref{fig:model}) learns video representations by creating representations of the video's metadata, and encouraging the video representations to match these metadata representations. The video representation is a vector $f_v \in \mathbb{R}^{D_v}$, and the metadata representation is a vector $f_t \in \mathbb{R}^{D_t}$, where the vector dimensions $D_v$ and $D_t$ are dependent on the models used to extract each representation and do not need to be the same.

Intuitively, the video and its metadata contain similar information, and therefore their representations $f_v$ and $f_t$ should contain similar information. However, the information contained in the video and its metadata are not exactly the same. The video will always contain information which is not present in the metadata. For example, the description of a rock climbing video will not list every hold the climber uses on their route. Likewise, the text will provide context which is not present in the video, such as listing the time and location where the video was shot. With our approach, we leverage this observation by encouraging the video representations to be similar, but not the same as, the corresponding metadata representation.

Specifically, the video representations (a 1-D vector for each video) are trained by \emph{predicting} the metadata representations (a 1-D vector for each piece of text). We predict the metadata representations from the video representations by applying a simple linear transformation, that is $\hat{f_t} = Wf_v + b$, where $W \in \mathbb{R}^{D_t \times D_v}$ and $b \in \mathbb{R}^{D_t}$. We then apply a ranking loss which penalizes $f_v$ if $\hat{f_t}$ is similar to the metadata representation for another video $f'_t$. That is,

\vspace{-3mm}

\begin{equation}
\label{eq:loss}
\mathcal{L}_{rank}(f_v, f_t, f'_t) = \mathrm{max}(0, m + d(\hat{f_t}, f_t) - d(\hat{f_t}, f'_t)),
\end{equation}

\noindent where $d$ is a distance metric, and $m$ is the minimum allowable margin between $d(\hat{f_t}, f_t)$ and $d(\hat{f_t}, f'_t)$. In our experiments, we set $d$ as the cosine distance, $d(u, v) = 1 - \frac{u^Tv}{\|u \|_2 \|v \|_2}$, and choose the margin to be $m = 0.1$.

For the loss, we require a ``negative'' metadata representation $f'_t$, that is, one drawn from a different video than $f_v$. We draw the negative example $f'_t$ from another video in the dataset uniformly at random. In addition, we use multiple negative examples $\{f'_{ti} \ | \ i = 1 \dots K\}$ for each positive example, and take the mean of their respective losses to get the final loss, $\mathcal{L} = \frac{1}{K} \sum_{i=1}^{K} \mathcal{L}_{rank}$. In practice, we use $K=15$, giving a ratio of 1 positive example for every 15 negative examples. We do not perform any hard-negative mining; we find that uniformly sampled negatives are sufficient. These negative examples are taken from the same batch of SGD training for convenience of implementation.

\noindent \textbf{Multiple Sources of Metadata}. When using more than one source of metadata for pre-training, we compute separate metadata representations $f_t$ for each source. Then, for each source, we apply a different set of linear transformation parameters $W, b$ to the video representation $f_v$, to compute a source-specific $\hat{f}_t$. We then separately compute a loss for each source as in Equation~\ref{eq:loss}. The final loss is the sum.

\noindent \textbf{End-to-End Video Representation Training}. We train the video representation $f_v$ end-to-end with the linear transformation parameters $W$ and $b$. Since our goal is to learn \emph{video} representations, not text representations, we do not train the metadata representations $f_t$ end-to-end. Instead, we use a pre-trained state-of-the-art text feature extractor to generate these embeddings (Section~\ref{sec:metadata}).

We train the model using stochastic gradient descent, with Nesterov momentum of 0.9~\cite{sutskever2013importance} and a weight decay of 1e-5. We apply dropout with a rate of 0.5 to the video features. We use a batch size of 2048 split into chunks of 16 videos across each of 128 accelerators, trained synchronously. The learning rate schedule begins with 1500 warmup steps for S3D-G or 2000 for R3D-50 (exponentially increasing from .001 to 1.0 for S3D-G and 1.6 for R3D-50), followed by a cosine-decaying~\cite{loshchilov2016sgdr} schedule for the remaining steps (140K for S3D-G, 120K for R3D-50). Due to the large batch size, training takes less than 3 days.

\subsection{Video Representation}
\label{sec:video}

We create the video representation $f_v \in \mathbb{R}^{D_v}$ using a 3D Convolutional Neural Network (3D CNN) which operates directly on the RGB video frames. The input to the 3D CNN is therefore a $H \times W \times T \times 3$ video clip. To get the video representation, we take the final hidden layer of the network and mean-pool across the spatial and temporal dimensions, resulting in a vector of length $D_v$.

In our work, we use S3D-G~\cite{xie2018rethinking} and 3D ResNet-50 (R3D-50)~\cite{he2016deep} as the backbone 3D CNN architectures. We choose these two architectures because R3D-50 provides high capacity and performance, while S3D-G offers lower computational cost while still outperforming well-known architectures such as I3D~\cite{carreira2017quo}. During training, we apply the 3D CNN on 64-frame clips drawn uniformly at random from the video at 25fps. For R3D-50 we sample 32 frames with stride 2 to match with ~\cite{qian2020spatiotemporal}. We resize the frames to 256px on the shortest edge, and then take a random $224\times224$ crop. We additionally perform brightness, contrast, and flipping augmentation. During inference, we use 250-frame clips (with circular padding where necessary), and take a center $224\times224$ crop. Similarly, we use 124 frames with stride 2 for R3D-50.

\subsection{Metadata Representation}
\label{sec:metadata}

For each piece of textual metadata, we create a metadata representation $f_{t} \in \mathbb{R}^{D_t}$ using BERT~\cite{devlin2018bert}, a state of the art text encoder. BERT returns a 768-dimensional embedding for each token in the text, and we take the mean of these token-level embeddings to get a single 768-dimensional representation of the metadata, that is, $D_t = 768$.

Specifically, we use the multilingual, cased version of BERT which was pre-trained on 104 languages, and has 12 layers and 110M parameters. We use the multilingual version because non-English text appears in WTS-70M. Since our goal is to learn \emph{video} representations, we do not fine-tune the BERT model. This also significantly alleviates the computational cost of training; otherwise fine-tuning the text model would dominate the computational cost.

When computing features for tags (where each video can have zero to many tags), we compute a BERT embedding for each individual tag and take the mean of the results. For videos with no tags, we replace it with an empty string. Each of the three other pieces of metadata (titles, descriptions, and channel names) are treated the same.

\section{Experiments}
\label{sec:experiments}

For some of our experiments, we use a subset of the full 70M-video dataset. These subsets are denoted by the approximate number of videos they include: 500K, 1M, 6M, 12M, 40M, and 70M. These subsets are not selected at random, instead each subset is chosen by selecting a smaller number of the top search results from each query, such that the 500K subset contains approximately the top 700 results per query and 70M contains 100K results per query. This reflects the way that such a method could be used in practice; one would search for queries relevant to their particular downstream task and collect as many of the top search results as they can, subject to space or bandwidth constraints.

We do not segment WTS-70M into a validation or test split, and instead evaluate our learned model purely by its performance on downstream tasks. We evaluate on three downstream video classification tasks:

\noindent \textbf{HMDB-51.} HMDB-51~\cite{kuehne2011hmdb} is an action recognition dataset consisting of short video clips associated with one of 51 classes. It contains 7000 videos, and is commonly used as a benchmark for video representation learning. We report results on the first test split, except where otherwise noted. When fine-tuning on HMDB-51, we use a learning rate of 1e-3 with an exponential decay schedule, a weight decay of 1e-7, and we train for 30 epochs.

\noindent \textbf{UCF-101.} UCF-101~\cite{soomro2012dataset} is a similar action recognition dataset consisting of video clips associated with one of 101 classes. It is larger than HMDB-51, consisting of over 13,000 videos. We report results on the first test split, except where otherwise noted. When fine-tuning on UCF-101, we use a learning rate of 1e-3 with an exponential decay schedule, a weight decay of 1e-7, and we train for 30 epochs.

\noindent \textbf{Kinetics-400, 600, 700.} Kinetics is a widely-used action recognition dataset consisting of 10-second clips drawn from videos annotated with action categories~\cite{kay2017kinetics}. Kinetics-400, 600, and 700 are increasingly large versions of the dataset, containing 400, 600, and 700 action categories, respectively~\cite{carreira2018short,carreira2019short}. Kinetics contains over 545K videos, and due to its scale, it is commonly used to pre-train video representations. We compare against Kinetics as a pre-training scheme, in addition to as a downstream task.

Kinetics videos can be deleted by their uploaders at any time, and then can no longer be recovered by researchers. Therefore, Kinetics gradually deteriorates over time, which generates discrepancies between both training and evaluation performed at different times. Our experiments were conducted using a snapshot of the Kinetics dataset collected in February 2020, when Kinetics-400 contained 225K of the original 247K training examples (-8.9\%), Kinetics-600 contained 378K of the original 393K training examples (-3.8\%), and Kinetics-700 contained 541K of the original 545K training examples (-0.7\%).

\subsection{Different Forms of Metadata}
\label{sec:metadataexps}

We collect four types of metadata for each video: the title, description, tags, and channel name (Section~\ref{sec:dataset}). We observe that each type of metadata contains a different level of detail and is affected by different sources of noise (Figure~\ref{fig:datasetexamples}). Therefore, we expect the different types of metadata to have different impacts on downstream performance. We investigate which of these are the most useful for pre-training in Table~\ref{tab:metadata}. For these experiments, we pre-train the model on WTS-500K and fine-tune on HMDB-51.

\setlength{\tabcolsep}{4pt}
\begin{table}[t]
\begin{center}
\begin{tabular}{lc}
\toprule
Supervision & HMDB-51 \\
\midrule [0.1em]
Scratch & 27.9 \\
\noalign{\smallskip}
\hline
\noalign{\smallskip}
Titles & 43.2 \\
Descriptions & 37.7 \\
Tags & 36.2 \\
Channel Name & 29.1\\
Titles + Desc. & 43.9\\
Titles + Desc. + Tags & 46.5 \\
All & \textbf{50.0} \\
\bottomrule [0.1em]
\end{tabular}
\vspace{2mm}
\caption{Sources of metadata used and their effect on downstream performance, as measured on HMDB-51 with an S3D-G backbone. Each source of metadata contributes individually to the final accuracy. For these experiments, we pre-train on WTS-500K. All reported accuracies are on HMDB-51 split 1.}
\label{tab:metadata}
\end{center}
\end{table}

We find that all types of metadata are useful sources of supervisory signal for pre-training. Titles are the most effective, achieving an increase in downstream accuracy of 15.3\% over a from-scratch baseline. Channel names are the least effective, resulting in only a 1.2\% improvement over the baseline. However, we find that these sources of supervision provide complementary signals, and that we achieve the best performance by including all of them during pre-training. This achieves a down-stream accuracy of 50.0\% on HMDB-51, a 22.1\% improvement over from-scratch.

In addition, these experiments can be used to show the relative utility of our pre-training and fully-supervised learning. These experiments are conducted using WTS-500K, which is approximately the same size as Kinetics-700 (545K videos). For comparison, a supervised model pre-trained on Kinetics-700 achieves 67.4\% accuracy on HMDB-51, a 17.4\% improvement over training on all four sources of metadata. As expected, weak textual supervision is not as effective, video for video, as supervised pre-training. However, this supervision does not require any labeling, making it an effective option.

\subsection{Scaling to 70M Videos}
\label{sec:scalingexps}

\begin{figure}[t]
\centering
\includegraphics[width=0.45\textwidth]{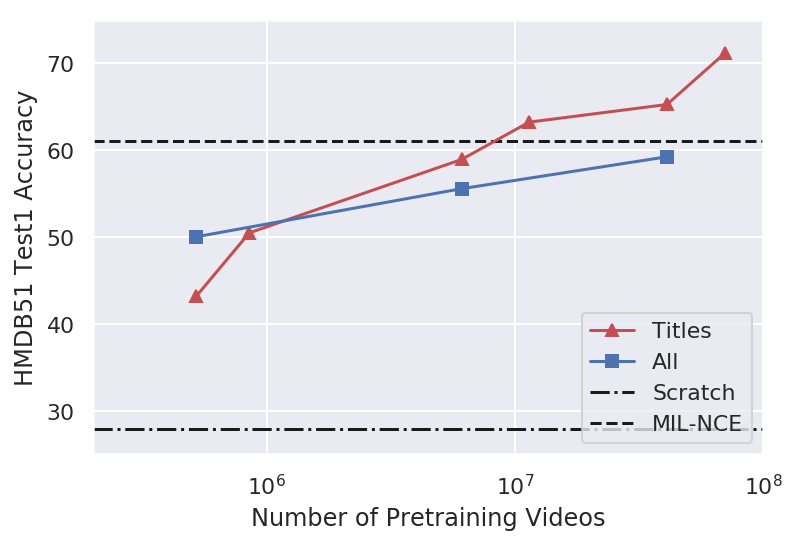}
\caption{Performance of our approach on HMDB-51 (split 1) with an S3D-G backbone for increasingly larger pre-training dataset sizes, compared to a baseline S3D-G model trained from scratch and one trained on HowTo100M with MIL-NCE~\cite{miech2019end}. Our approach can learn effectively from 70M videos and exceeds the performance of MIL-NCE with 12M videos.}
\label{fig:scaling}
\end{figure}

To demonstrate the scalability of our method, we apply it to increasingly large subsets of the full 70M-video dataset in Figure~\ref{fig:scaling}. We compare two metadata configurations for this experiment: (1) only titles, and (2) all metadata. We find that the titles-only approach scales significantly better than the all-metadata approach; although using all metadata leads to higher downstream accuracy with 500K pre-training videos, this is reversed when using more than 1M pre-training videos. This is likely due to the poor scaling properties of tags and descriptions as shown in Figure~\ref{fig:scalinganalysis}, and suggests that noise can become a burden on training.

For the titles-only approach, we find that using more pre-training data sharply improves performance. Using all 70M videos for pre\--training achieves an HMDB-51 accuracy of 71.1\%, a 13.2\% improvement over using 500K videos. In addition, this accuracy exceeds that of an equivalent model trained on Kinetics\--700 (67.4\%), demonstrating that our approach can exceed the performance of fully\--supervised pre\--training, without any labeling cost. We also note that performance does not appear to have saturated by the time we hit 70M videos, indicating that future work could demonstrate even further gains by collecting more videos.

For these results, we only use S3D-G and do not expand the model capacity when scaling to 70M videos. Interestingly, we achieve good performance on the 70M dataset using only 4 epochs of training, while on the 500K dataset we require over 80 epochs of training. This suggests that increased model capacity could further improve performance for the 70M dataset. In the following section, we improve these results further using a R3D-50 architecture.

\setlength{\tabcolsep}{1.5pt}
\begin{table}[t]
\footnotesize
\begin{center}
\begin{tabular}{llllccc}
\toprule
Method & Data & Mod. & Model & Frozen & HMDB & UCF \\
\midrule [0.1em]
Baseline & N/A & N/A & S3D-G & & 27.9 & 58.5\\
 \noalign{\smallskip}
\hline
\noalign{\smallskip}
DPC~\cite{han2019video} & K400 & RGB & R3D-34 & & 35.7 & 75.7 \\
S\&L~\cite{misra2016shuffle}* & K600 & RGB & S3D & & 35.8 & 68.7 \\ 
3DRotNet~\cite{jing2018self}* & K600 & RGB & S3D & &  40.0 & 75.3 \\
CBT~\cite{sun2019contrastive} & K600 & RGB & S3D & &  44.6 & 79.5 \\
VTHCL~\cite{yang2020video} & K400 & RGB & R3D-50 & & 49.2 & 82.1 \\
AVTS~\cite{korbar2018cooperative} & K600 & Audio & I3D & & 53.0 & 83.7 \\
CPD~\cite{li2020learning} & K400 & Titles & R3D-34 & & 57.7 & 88.7 \\
MIL-NCE~\cite{miech2019end} & HT100M & Speech & S3D-G & \checkmark & 53.1 & 82.7 \\
MIL-NCE~\cite{miech2019end} & HT100M & Speech & S3D-G & & 61.0 & 91.3 \\
XDC~\cite{alwassel2019self} & IG65M & Audio & R3D-18 & & 63.1 & 91.5 \\
CVRL~\cite{qian2020spatiotemporal} & K400 & RGB & R3D-50 & \checkmark & 57.3 & 89.2 \\
CVRL~\cite{qian2020spatiotemporal} & K400 & RGB & R3D-50 & & 66.7 & 92.2 \\
CVRL~\cite{qian2020spatiotemporal} & K600 & RGB & R3D-50 & \checkmark & 59.7 & 90.6 \\
CVRL~\cite{qian2020spatiotemporal} & K600 & RGB & R3D-50 & & 68.0 & 93.4 \\
\noalign{\smallskip}
\hline
\noalign{\smallskip}
Ours & WTS-70M & Titles & S3D-G & \checkmark & 56.5 & 84.5 \\
Ours & WTS-70M & Titles & S3D-G & & 71.1 & 92.7 \\
Ours & WTS-70M & Titles & R3D-50 & \checkmark & \textbf{70.3} & \textbf{94.7} \\
Ours & WTS-70M & Titles & R3D-50 & & \textbf{77.7} & \textbf{95.8} \\ 
\bottomrule [0.1em]
\end{tabular}
\vspace{2mm}
\caption{Comparison with self-supervised pre-training prior work on HMDB-51 and UCF-101. ``Data'' refers to the source of pre-training videos, however, these approaches do not use the available labels. ''Mod.'' refers to the modalities used to train each model. All numbers are quoted directly from the original authors. Our results are averaged across all three splits of HMDB-51 and UCF-101. *Reimplemented by~\cite{sun2019contrastive}.}
\label{table:compare}
\end{center}
\end{table}
\setlength{\tabcolsep}{4pt}

\subsection{Comparison with Prior Work}
\label{sec:comparisonexps}

In Table~\ref{table:compare}, we compare our approach against other methods for self-supervised and webly-supervised learning. We strongly outperform all existing methods, suggesting that the textual metadata provides a strong supervisory signal compared to these methods. Notably, when using the same S3D-G backbone, we outperform MIL-NCE~\cite{miech2019end}, a recent method for learning video representations from instructional videos in the HowTo100M dataset~\cite{miech2019howto100m}, both on HMDB-51 (+10.1\%) and UCF101 (+1.4\%).

We additionally train R3D-50, a much higher-capacity model than S3D-G, for 20 epochs on WTS-70M to achieve state of the art results. Specifically, we achieve 76.9\% on HMDB-51 and 95.5\% on UCF101 \emph{without any labeled data during pre-training}, beating the previous best results from CVRL~\cite{qian2020spatiotemporal} by 8.9\% on HMDB-51 and 2.1\% on UCF101. We also strongly outperform the results from MIL-NCE and XDC, which are trained using cross-modal self-supervised learning approaches on large-scale datasets, similar to our approach. This suggests that titles provide rich supervision compared to these approaches, which use speech and audio.

We additionally report results using frozen feature vectors, as opposed to fully fine-tuned networks (non-frozen). This experiment represents the common real-world case in which full fine-tuning is not possible due to multiple tasks sharing the same features, or not having sufficient compute to fully fine-tune the network. In this case, we find that WTS still outperforms prior work that reports these results, specifically MIL-NCE using the S3D-G backbone.

In Table~\ref{table:kinetics}, we present results on Kinetics-400 and Kinetics-600. For these results, we pre-train on WTS-70M and either freeze or fully finetune the backbone network, adding one linear layer to classify the Kinetics videos. We find that our pre-training improves performance over a model trained from scratch on both Kinetics-400 and 600, demonstrating that WTS can also be used with large labeled datasets. We also find that WTS outperforms CVRL when training from frozen embeddings on Kinetics-400~\cite{qian2020spatiotemporal}, though performs slightly worse on Kinetics-600. Note that CVRL assumes access to the Kinetics dataset (though unlabeled) during pre-training, while our model does not.

\begin{table}[t]
\begin{center}
\begin{tabular}{lcccc}
\toprule
Method & Model & Frozen & K400 & K600 \\
\midrule [0.1em]
VINCE~\cite{gordon2020watching} & R3D-50 & \checkmark & 36.2 & - \\
VTHCL~\cite{yang2020video} & R3D-50 & \checkmark & 37.8 & - \\
CVRL~\cite{qian2020spatiotemporal} & R3D-50 & \checkmark & 66.1 & 70.4 \\
WTS-70M (Ours) & R3D-50 & \checkmark & \textbf{72.7} & \textbf{75.5}\\
\hline\noalign{\smallskip}
Scratch & R3D-50 & & 73.4 & 78.8 \\
WTS-70M (Ours) & R3D-50 & & \textbf{75.1} & \textbf{79.3} \\
\bottomrule [0.1em]
\end{tabular}
\vspace{2mm}
\caption{Experiments on Kinetics. Our method and baselines are trained without labels and we report classification performance for both frozen feature and fine-tuning experiments.}
\label{table:kinetics}
\end{center}
\vspace{-3mm}
\end{table}

\subsection{Semi-Supervised Learning}
\label{sec:fewshot}

\begin{table}[t]
\begin{center}
\begin{tabular}{lccc}
\toprule
 & & \multicolumn{2}{c}{Label fraction} \\
Method & Model & 1\% & 10\% \\
\midrule [0.1em]
Scratch & R3D-50 & 4.3 & 45.3 \\
\hline\noalign{\smallskip}
CVRL~\cite{qian2020spatiotemporal} & R3D-50 & 36.7 & 56.1 \\
\hline\noalign{\smallskip}
WTS-70M (Ours) & R3D-50 & \textbf{40.9} & \textbf{61.5} \\
\bottomrule [0.1em]
\end{tabular}
\vspace{2mm}
\caption{Comparison with semi-supervised learning on Kinetics-600. We finetune WTS using a fraction of the labeled videos in Kinetics-600. CVRL~\cite{qian2020spatiotemporal} uses unlabeled Kinetics-600 videos for pre-training, which makes their approach semi-supervised.}
\label{tab:semisupervised}
\end{center}
\end{table}

We additionally perform an experiment in the extreme low-data regime. Specifically, we use random subsets of Kinetics-600 to fine-tune a WTS pre-trained R3D-50 model. These subsets have as few as 6.5 examples per class on average, making them much more challenging than the full Kinetics-600 dataset. In Table~\ref{tab:semisupervised}, we show that WTS-70M pre-training strongly outperforms from-scratch training, and also outperforms CVRL~\cite{qian2020spatiotemporal}.

\subsection{Complementary Strong and Weak Supervision}
\label{sec:complementaryexps}

\begin{table}[t]
\begin{center}
\begin{tabular}{lcc}
\toprule
Pre-training & Model & HMDB-51 \\
\midrule [0.1em]
70M & S3D-G & 67.4 \\
K700 & S3D-G & 71.1 \\\noalign{\smallskip} \hline\noalign{\smallskip}
70M+K400 & S3D-G & 72.2 \\
70M+K600 & S3D-G & 74.5 \\
70M+K700 & S3D-G & \textbf{75.9} \\
\bottomrule [0.1em]
\end{tabular}
\vspace{2mm}
\caption{Complementary nature of our approach and fully-supervised learning (using S3D-G backbone). We pre-train the model on WTS-70M, then fine-tune it on Kinetics, then apply it to HMDB-51 (split 1). K\emph{X} = Kinetics-\emph{X}.}
\label{table:complementary}
\end{center}
\vspace{-3mm}
\end{table}

Our approach learning has the capacity to exceed the performance of strongly-supervised learning, without any labels (Section~\ref{sec:scalingexps}). However, in practice, one would use all sources of supervision available, including labeled datasets. Therefore, we ask whether our approach and strongly-supervised learning can be applied in combination, to further improve the performance on down-stream tasks. We test this in Table~\ref{table:complementary} by training in a three-step process: first, we pre-train our model on WTS-70M. Then, we fine-tune this model on Kinetics. Finally, we apply the resulting model to HMDB-51.

We find that our approach and strongly-supervised learning are indeed complementary. When using both WTS-70M and Kinetics-700 in combination, the down-stream accuracy on HMDB-51 increases by a further 8.5\% over Kinetics-700 alone. This demonstrates that our method is effective even in situations where labeled data is already plentiful.

\section{Conclusions}

We demonstrate that textual metadata serves as a useful signal for pre-training video representations, without the need for any manually annotated labels. Specifically, we find that each textual signal is complementary (Section~\ref{sec:metadataexps}), and that this approach exceeds the performance of supervised pre\--training when scaled to tens of millions of videos (Section~\ref{sec:scalingexps}). We also show that it outperforms competitive approaches for both self-supervised and webly-supervised learning (Section~\ref{sec:comparisonexps}). Finally, we demonstrate that it works in extreme low-data regimes (Section~\ref{sec:fewshot}) and is complementary with full supervision (Section~\ref{sec:complementaryexps}). These findings suggest that textual metadata can be used as an effective pre-training strategy for a wide variety of downstream tasks.

{\small
\bibliographystyle{cvpr/ieee_fullname}
\bibliography{egbib}

\begin{thebibliography}{10}\itemsep=-1pt

\bibitem{abu2016youtube}
Sami Abu-El-Haija, Nisarg Kothari, Joonseok Lee, Paul Natsev, George Toderici,
  Balakrishnan Varadarajan, and Sudheendra Vijayanarasimhan.
\newblock Youtube-8m: A large-scale video classification benchmark.
\newblock {\em arXiv preprint arXiv:1609.08675}, 2016.

\bibitem{alayrac2016unsupervised}
Jean-Baptiste Alayrac, Piotr Bojanowski, Nishant Agrawal, Josef Sivic, Ivan
  Laptev, and Simon Lacoste-Julien.
\newblock Unsupervised learning from narrated instruction videos.
\newblock In {\em Proceedings of the IEEE Conference on Computer Vision and
  Pattern Recognition}, pages 4575--4583, 2016.

\bibitem{alwassel2019self}
Humam Alwassel, Dhruv Mahajan, Lorenzo Torresani, Bernard Ghanem, and Du Tran.
\newblock Self-supervised learning by cross-modal audio-video clustering.
\newblock {\em arXiv preprint arXiv:1911.12667}, 2019.

\bibitem{arandjelovic2017look}
Relja Arandjelovic and Andrew Zisserman.
\newblock Look, listen and learn.
\newblock In {\em Proceedings of the IEEE International Conference on Computer
  Vision}, pages 609--617, 2017.

\bibitem{aytar2016soundnet}
Yusuf Aytar, Carl Vondrick, and Antonio Torralba.
\newblock Soundnet: Learning sound representations from unlabeled video.
\newblock In {\em Advances in neural information processing systems}, pages
  892--900, 2016.

\bibitem{bergamo2010exploiting}
Alessandro Bergamo and Lorenzo Torresani.
\newblock Exploiting weakly-labeled web images to improve object
  classification: a domain adaptation approach.
\newblock In {\em Advances in neural information processing systems}, pages
  181--189, 2010.

\bibitem{carreira2018short}
Joao Carreira, Eric Noland, Andras Banki-Horvath, Chloe Hillier, and Andrew
  Zisserman.
\newblock A short note about kinetics-600.
\newblock {\em arXiv preprint arXiv:1808.01340}, 2018.

\bibitem{carreira2019short}
Joao Carreira, Eric Noland, Chloe Hillier, and Andrew Zisserman.
\newblock A short note on the kinetics-700 human action dataset.
\newblock {\em arXiv preprint arXiv:1907.06987}, 2019.

\bibitem{carreira2017quo}
Joao Carreira and Andrew Zisserman.
\newblock Quo vadis, action recognition? a new model and the kinetics dataset.
\newblock In {\em proceedings of the IEEE Conference on Computer Vision and
  Pattern Recognition}, pages 6299--6308, 2017.

\bibitem{chen2015webly}
Xinlei Chen and Abhinav Gupta.
\newblock Webly supervised learning of convolutional networks.
\newblock In {\em Proceedings of the IEEE International Conference on Computer
  Vision}, pages 1431--1439, 2015.

\bibitem{chen2013neil}
Xinlei Chen, Abhinav Shrivastava, and Abhinav Gupta.
\newblock Neil: Extracting visual knowledge from web data.
\newblock In {\em Proceedings of the IEEE International Conference on Computer
  Vision}, pages 1409--1416, 2013.

\bibitem{devlin2018bert}
Jacob Devlin, Ming-Wei Chang, Kenton Lee, and Kristina Toutanova.
\newblock Bert: Pre-training of deep bidirectional transformers for language
  understanding.
\newblock {\em arXiv preprint arXiv:1810.04805}, 2018.

\bibitem{divvala2014learning}
Santosh~K Divvala, Ali Farhadi, and Carlos Guestrin.
\newblock Learning everything about anything: Webly-supervised visual concept
  learning.
\newblock In {\em Proceedings of the IEEE Conference on Computer Vision and
  Pattern Recognition}, pages 3270--3277, 2014.

\bibitem{dwibedi2019temporal}
Debidatta Dwibedi, Yusuf Aytar, Jonathan Tompson, Pierre Sermanet, and Andrew
  Zisserman.
\newblock Temporal cycle-consistency learning.
\newblock In {\em Proceedings of the IEEE Conference on Computer Vision and
  Pattern Recognition}, pages 1801--1810, 2019.

\bibitem{fergus2010learning}
Rob Fergus, Li Fei-Fei, Pietro Perona, and Andrew Zisserman.
\newblock Learning object categories from internet image searches.
\newblock {\em Proceedings of the IEEE}, 98(8):1453--1466, 2010.

\bibitem{fernando2017self}
Basura Fernando, Hakan Bilen, Efstratios Gavves, and Stephen Gould.
\newblock Self-supervised video representation learning with odd-one-out
  networks.
\newblock In {\em Proceedings of the IEEE conference on computer vision and
  pattern recognition}, pages 3636--3645, 2017.

\bibitem{gan2018geometry}
Chuang Gan, Boqing Gong, Kun Liu, Hao Su, and Leonidas~J Guibas.
\newblock Geometry guided convolutional neural networks for self-supervised
  video representation learning.
\newblock In {\em Proceedings of the IEEE Conference on Computer Vision and
  Pattern Recognition}, pages 5589--5597, 2018.

\bibitem{gan2016webly}
Chuang Gan, Chen Sun, Lixin Duan, and Boqing Gong.
\newblock Webly-supervised video recognition by mutually voting for relevant
  web images and web video frames.
\newblock In {\em European Conference on Computer Vision}, pages 849--866.
  Springer, 2016.

\bibitem{ghadiyaram2019large}
Deepti Ghadiyaram, Du Tran, and Dhruv Mahajan.
\newblock Large-scale weakly-supervised pre-training for video action
  recognition.
\newblock In {\em Proceedings of the IEEE Conference on Computer Vision and
  Pattern Recognition}, pages 12046--12055, 2019.

\bibitem{gong2014multi}
Yunchao Gong, Qifa Ke, Michael Isard, and Svetlana Lazebnik.
\newblock A multi-view embedding space for modeling internet images, tags, and
  their semantics.
\newblock {\em International journal of computer vision}, 106(2):210--233,
  2014.

\bibitem{gong2014improving}
Yunchao Gong, Liwei Wang, Micah Hodosh, Julia Hockenmaier, and Svetlana
  Lazebnik.
\newblock Improving image-sentence embeddings using large weakly annotated
  photo collections.
\newblock In {\em European conference on computer vision}, pages 529--545.
  Springer, 2014.

\bibitem{gordon2020watching}
Daniel Gordon, Kiana Ehsani, Dieter Fox, and Ali Farhadi.
\newblock Watching the world go by: Representation learning from unlabeled
  videos.
\newblock {\em arXiv preprint arXiv:2003.07990}, 2020.

\bibitem{gu2018ava}
Chunhui Gu, Chen Sun, David~A Ross, Carl Vondrick, Caroline Pantofaru, Yeqing
  Li, Sudheendra Vijayanarasimhan, George Toderici, Susanna Ricco, Rahul
  Sukthankar, et~al.
\newblock Ava: A video dataset of spatio-temporally localized atomic visual
  actions.
\newblock In {\em Proceedings of the IEEE Conference on Computer Vision and
  Pattern Recognition}, pages 6047--6056, 2018.

\bibitem{tubefilter2019}
James Hale.
\newblock {More Than 500 Hours Of Content Are Now Being Uploaded To YouTube
  Every Minute}, 2019.

\bibitem{han2019video}
Tengda Han, Weidi Xie, and Andrew Zisserman.
\newblock Video representation learning by dense predictive coding.
\newblock In {\em Proceedings of the IEEE International Conference on Computer
  Vision Workshops}, pages 0--0, 2019.

\bibitem{he2016deep}
Kaiming He, Xiangyu Zhang, Shaoqing Ren, and Jian Sun.
\newblock Deep residual learning for image recognition.
\newblock In {\em Proceedings of the IEEE conference on computer vision and
  pattern recognition}, pages 770--778, 2016.

\bibitem{allegro2019}
Gal Hyams, Dan Malowany, Ariel Biller, and Gregory Axler.
\newblock {Quantifying Diminishing Returns of Annotated Data}, 2019.

\bibitem{izadinia2015deep}
Hamid Izadinia, Bryan~C Russell, Ali Farhadi, Matthew~D Hoffman, and Aaron
  Hertzmann.
\newblock Deep classifiers from image tags in the wild.
\newblock In {\em Proceedings of the 2015 Workshop on Community-Organized
  Multimodal Mining: Opportunities for Novel Solutions}, pages 13--18, 2015.

\bibitem{jing2018self}
Longlong Jing and Yingli Tian.
\newblock Self-supervised spatiotemporal feature learning by video geometric
  transformations.
\newblock {\em arXiv preprint arXiv:1811.11387}, 2(7):8, 2018.

\bibitem{joulin2016learning}
Armand Joulin, Laurens van~der Maaten, Allan Jabri, and Nicolas Vasilache.
\newblock Learning visual features from large weakly supervised data.
\newblock In {\em European Conference on Computer Vision}, pages 67--84.
  Springer, 2016.

\bibitem{karpathy2014large}
Andrej Karpathy, George Toderici, Sanketh Shetty, Thomas Leung, Rahul
  Sukthankar, and Li Fei-Fei.
\newblock Large-scale video classification with convolutional neural networks.
\newblock In {\em Proceedings of the IEEE conference on Computer Vision and
  Pattern Recognition}, pages 1725--1732, 2014.

\bibitem{kay2017kinetics}
Will Kay, Joao Carreira, Karen Simonyan, Brian Zhang, Chloe Hillier, Sudheendra
  Vijayanarasimhan, Fabio Viola, Tim Green, Trevor Back, Paul Natsev, et~al.
\newblock The kinetics human action video dataset.
\newblock {\em arXiv preprint arXiv:1705.06950}, 2017.

\bibitem{kim2019self}
Dahun Kim, Donghyeon Cho, and In~So Kweon.
\newblock Self-supervised video representation learning with space-time cubic
  puzzles.
\newblock In {\em Proceedings of the AAAI Conference on Artificial
  Intelligence}, volume~33, pages 8545--8552, 2019.

\bibitem{korbar2018cooperative}
Bruno Korbar, Du Tran, and Lorenzo Torresani.
\newblock Cooperative learning of audio and video models from self-supervised
  synchronization.
\newblock In {\em Advances in Neural Information Processing Systems}, pages
  7763--7774, 2018.

\bibitem{kuehne2019mining}
Hilde Kuehne, Ahsan Iqbal, Alexander Richard, and Juergen Gall.
\newblock Mining youtube-a dataset for learning fine-grained action concepts
  from webly supervised video data.
\newblock {\em arXiv preprint arXiv:1906.01012}, 2019.

\bibitem{kuehne2011hmdb}
Hildegard Kuehne, Hueihan Jhuang, Est{\'\i}baliz Garrote, Tomaso Poggio, and
  Thomas Serre.
\newblock Hmdb: a large video database for human motion recognition.
\newblock In {\em 2011 International Conference on Computer Vision}, pages
  2556--2563. IEEE, 2011.

\bibitem{lai2019self}
Zihang Lai and Weidi Xie.
\newblock Self-supervised learning for video correspondence flow.
\newblock {\em arXiv preprint arXiv:1905.00875}, 2019.

\bibitem{lee2017unsupervised}
Hsin-Ying Lee, Jia-Bin Huang, Maneesh Singh, and Ming-Hsuan Yang.
\newblock Unsupervised representation learning by sorting sequences.
\newblock In {\em Proceedings of the IEEE International Conference on Computer
  Vision}, pages 667--676, 2017.

\bibitem{li2017learning}
Ang Li, Allan Jabri, Armand Joulin, and Laurens van~der Maaten.
\newblock Learning visual n-grams from web data.
\newblock In {\em Proceedings of the IEEE International Conference on Computer
  Vision}, pages 4183--4192, 2017.

\bibitem{li2020learning}
Tianhao Li and Limin Wang.
\newblock Learning spatiotemporal features via video and text pair
  discrimination.
\newblock {\em arXiv preprint arXiv:2001.05691}, 2020.

\bibitem{loshchilov2016sgdr}
Ilya Loshchilov and Frank Hutter.
\newblock Sgdr: Stochastic gradient descent with warm restarts.
\newblock {\em arXiv preprint arXiv:1608.03983}, 2016.

\bibitem{lotter2016deep}
William Lotter, Gabriel Kreiman, and David Cox.
\newblock Deep predictive coding networks for video prediction and unsupervised
  learning.
\newblock {\em arXiv preprint arXiv:1605.08104}, 2016.

\bibitem{mahajan2018exploring}
Dhruv Mahajan, Ross Girshick, Vignesh Ramanathan, Kaiming He, Manohar Paluri,
  Yixuan Li, Ashwin Bharambe, and Laurens van~der Maaten.
\newblock Exploring the limits of weakly supervised pretraining.
\newblock In {\em Proceedings of the European Conference on Computer Vision
  (ECCV)}, pages 181--196, 2018.

\bibitem{mathieu2015deep}
Michael Mathieu, Camille Couprie, and Yann LeCun.
\newblock Deep multi-scale video prediction beyond mean square error.
\newblock {\em arXiv preprint arXiv:1511.05440}, 2015.

\bibitem{miech2019end}
Antoine Miech, Jean-Baptiste Alayrac, Lucas Smaira, Ivan Laptev, Josef Sivic,
  and Andrew Zisserman.
\newblock End-to-end learning of visual representations from uncurated
  instructional videos.
\newblock {\em arXiv preprint arXiv:1912.06430}, 2019.

\bibitem{miech2019howto100m}
Antoine Miech, Dimitri Zhukov, Jean-Baptiste Alayrac, Makarand Tapaswi, Ivan
  Laptev, and Josef Sivic.
\newblock Howto100m: Learning a text-video embedding by watching hundred
  million narrated video clips.
\newblock In {\em Proceedings of the IEEE International Conference on Computer
  Vision}, pages 2630--2640, 2019.

\bibitem{misra2016shuffle}
Ishan Misra, C~Lawrence Zitnick, and Martial Hebert.
\newblock Shuffle and learn: unsupervised learning using temporal order
  verification.
\newblock In {\em European Conference on Computer Vision}, pages 527--544.
  Springer, 2016.

\bibitem{mithun2018webly}
Niluthpol~Chowdhury Mithun, Rameswar Panda, Evangelos~E Papalexakis, and Amit~K
  Roy-Chowdhury.
\newblock Webly supervised joint embedding for cross-modal image-text
  retrieval.
\newblock In {\em Proceedings of the 26th ACM international conference on
  Multimedia}, pages 1856--1864, 2018.

\bibitem{nagrani_cvpr_2020}
Arsha Nagrani, Sun Chen, David Ross, Rahul Sukthankar, Cordelia Schmid, and
  Andrew Zisserman.
\newblock Speech2action: Cross-modal supervision for action recognition.
\newblock {\em CVPR}, 2020.

\bibitem{ordonez2011im2text}
Vicente Ordonez, Girish Kulkarni, and Tamara~L Berg.
\newblock Im2text: Describing images using 1 million captioned photographs.
\newblock In {\em Advances in neural information processing systems}, pages
  1143--1151, 2011.

\bibitem{owens2018audio}
Andrew Owens and Alexei~A Efros.
\newblock Audio-visual scene analysis with self-supervised multisensory
  features.
\newblock In {\em Proceedings of the European Conference on Computer Vision
  (ECCV)}, pages 631--648, 2018.

\bibitem{owens2016ambient}
Andrew Owens, Jiajun Wu, Josh~H McDermott, William~T Freeman, and Antonio
  Torralba.
\newblock Ambient sound provides supervision for visual learning.
\newblock In {\em European conference on computer vision}, pages 801--816.
  Springer, 2016.

\bibitem{pathak2017learning}
Deepak Pathak, Ross Girshick, Piotr Doll{\'a}r, Trevor Darrell, and Bharath
  Hariharan.
\newblock Learning features by watching objects move.
\newblock In {\em Proceedings of the IEEE Conference on Computer Vision and
  Pattern Recognition}, pages 2701--2710, 2017.

\bibitem{qian2020spatiotemporal}
Rui Qian, Tianjian Meng, Boqing Gong, Ming-Hsuan Yang, Huisheng Wang, Serge
  Belongie, and Yin Cui.
\newblock Spatiotemporal contrastive video representation learning.
\newblock {\em arXiv preprint arXiv:2008.03800}, 2020.

\bibitem{rouditchenko2019self}
Andrew Rouditchenko, Hang Zhao, Chuang Gan, Josh McDermott, and Antonio
  Torralba.
\newblock Self-supervised audio-visual co-segmentation.
\newblock In {\em ICASSP 2019-2019 IEEE International Conference on Acoustics,
  Speech and Signal Processing (ICASSP)}, pages 2357--2361. IEEE, 2019.

\bibitem{schroff2010harvesting}
Florian Schroff, Antonio Criminisi, and Andrew Zisserman.
\newblock Harvesting image databases from the web.
\newblock {\em IEEE transactions on pattern analysis and machine intelligence},
  33(4):754--766, 2010.

\bibitem{soomro2012dataset}
Khurram Soomro, Amir~Roshan Zamir, and M Shah.
\newblock A dataset of 101 human action classes from videos in the wild.
\newblock {\em Center for Research in Computer Vision}, 2, 2012.

\bibitem{sun2019contrastive}
Chen Sun, Fabien Baradel, Kevin Murphy, and Cordelia Schmid.
\newblock Contrastive bidirectional transformer for temporal representation
  learning.
\newblock {\em arXiv preprint arXiv:1906.05743}, 2019.

\bibitem{sutskever2013importance}
Ilya Sutskever, James Martens, George Dahl, and Geoffrey Hinton.
\newblock On the importance of initialization and momentum in deep learning.
\newblock In {\em International conference on machine learning}, pages
  1139--1147, 2013.

\bibitem{vondrick2016anticipating}
Carl Vondrick, Hamed Pirsiavash, and Antonio Torralba.
\newblock Anticipating visual representations from unlabeled video.
\newblock In {\em Proceedings of the IEEE Conference on Computer Vision and
  Pattern Recognition}, pages 98--106, 2016.

\bibitem{vondrick2016generating}
Carl Vondrick, Hamed Pirsiavash, and Antonio Torralba.
\newblock Generating videos with scene dynamics.
\newblock In {\em Advances in neural information processing systems}, pages
  613--621, 2016.

\bibitem{vondrick2018tracking}
Carl Vondrick, Abhinav Shrivastava, Alireza Fathi, Sergio Guadarrama, and Kevin
  Murphy.
\newblock Tracking emerges by colorizing videos.
\newblock In {\em Proceedings of the European Conference on Computer Vision
  (ECCV)}, pages 391--408, 2018.

\bibitem{wang2019self}
Jiangliu Wang, Jianbo Jiao, Linchao Bao, Shengfeng He, Yunhui Liu, and Wei Liu.
\newblock Self-supervised spatio-temporal representation learning for videos by
  predicting motion and appearance statistics.
\newblock In {\em Proceedings of the IEEE Conference on Computer Vision and
  Pattern Recognition}, pages 4006--4015, 2019.

\bibitem{wang2019learning}
Xiaolong Wang, Allan Jabri, and Alexei~A Efros.
\newblock Learning correspondence from the cycle-consistency of time.
\newblock In {\em Proceedings of the IEEE Conference on Computer Vision and
  Pattern Recognition}, pages 2566--2576, 2019.

\bibitem{wang2008annotating}
Xin-Jing Wang, Lei Zhang, Xirong Li, and Wei-Ying Ma.
\newblock Annotating images by mining image search results.
\newblock {\em IEEE Transactions on Pattern Analysis and Machine Intelligence},
  30(11):1919--1932, 2008.

\bibitem{wei2018learning}
Donglai Wei, Joseph~J Lim, Andrew Zisserman, and William~T Freeman.
\newblock Learning and using the arrow of time.
\newblock In {\em Proceedings of the IEEE Conference on Computer Vision and
  Pattern Recognition}, pages 8052--8060, 2018.

\bibitem{xie2018rethinking}
Saining Xie, Chen Sun, Jonathan Huang, Zhuowen Tu, and Kevin Murphy.
\newblock Rethinking spatiotemporal feature learning: Speed-accuracy trade-offs
  in video classification.
\newblock In {\em Proceedings of the European Conference on Computer Vision
  (ECCV)}, pages 305--321, 2018.

\bibitem{xu2019self}
Dejing Xu, Jun Xiao, Zhou Zhao, Jian Shao, Di Xie, and Yueting Zhuang.
\newblock Self-supervised spatiotemporal learning via video clip order
  prediction.
\newblock In {\em Proceedings of the IEEE Conference on Computer Vision and
  Pattern Recognition}, pages 10334--10343, 2019.

\bibitem{yang2020video}
Ceyuan Yang, Yinghao Xu, Bo Dai, and Bolei Zhou.
\newblock Video representation learning with visual tempo consistency.
\newblock {\em arXiv preprint arXiv:2006.15489}, 2020.

\bibitem{yu2014instructional}
Shoou-I Yu, Lu Jiang, and Alexander Hauptmann.
\newblock Instructional videos for unsupervised harvesting and learning of
  action examples.
\newblock In {\em Proceedings of the 22nd ACM international conference on
  Multimedia}, pages 825--828, 2014.

\bibitem{zhao2018sound}
Hang Zhao, Chuang Gan, Andrew Rouditchenko, Carl Vondrick, Josh McDermott, and
  Antonio Torralba.
\newblock The sound of pixels.
\newblock In {\em Proceedings of the European Conference on Computer Vision
  (ECCV)}, pages 570--586, 2018.

\bibitem{zhou2018towards}
Luowei Zhou, Chenliang Xu, and Jason~J Corso.
\newblock Towards automatic learning of procedures from web instructional
  videos.
\newblock In {\em Thirty-Second AAAI Conference on Artificial Intelligence},
  2018.

\bibitem{zhukov2019cross}
Dimitri Zhukov, Jean-Baptiste Alayrac, Ramazan~Gokberk Cinbis, David Fouhey,
  Ivan Laptev, and Josef Sivic.
\newblock Cross-task weakly supervised learning from instructional videos.
\newblock In {\em Proceedings of the IEEE Conference on Computer Vision and
  Pattern Recognition}, pages 3537--3545, 2019.

\end{thebibliography}
}

\newpage

\begin{appendices}

\section{Scaling to 70M Details}

In Table~\ref{tab:scaling}, we show the number of pre-training iterations used for the scaling experiments (Fig. 4 in the main paper). Other than changing the number of pre-training iterations, we do not modify any hyper-parameters when training on the smaller subsets of WTS-70M.

\begin{table}[h]
\centering
\begin{tabular}{lcc}
\toprule
Dataset & Iters & HMDB-51 \\
\midrule [0.1em]
Scratch & N/A & 27.9 \\
\noalign{\smallskip}
\hline
\noalign{\smallskip}
500K & 20K & 43.2 \\
1M & 25K & 50.5 \\
6M & 30K & 58.9 \\
12M & 50K & 63.2 \\
40M & 100K & 65.2 \\
70M & 120K & \textbf{71.1} \\
\noalign{\smallskip}
\hline
\noalign{\smallskip}
\textcolor{lightgray}{K700} & \textcolor{lightgray}{30K} & \textcolor{lightgray}{67.4} \\
\bottomrule [0.1em]
\vspace{2mm}
\end{tabular}
\caption{Number of pre-training iterations and resulting accuracy for scaling experiments with S3D-G backbone. K700 = fully supervised pretraining on Kinetics-700 (not comparable).} 
\label{tab:scaling}
\end{table}

\section{Semi-Supervised Learning}

In Table~\ref{tab:fewshot}, we show extended results from our semi-supervised learning experiments (Table~\ref{tab:semisupervised}).

\begin{table}[h]
\centering
\begin{tabular}{rcc}
\hline\noalign{\smallskip}
Videos Used & Scratch & WTS-70M \\
\noalign{\smallskip}
\hline
\noalign{\smallskip}
1\% & 6.0 & 40.9 \\
2\% & 14.1 & 47.7 \\
5\% & 22.5 & 56.7 \\
10\% & 47.4 & 61.5 \\
25\% & 58.0 & 65.4 \\
50\% & 72.8 & 74.0 \\
100\% & 78.8 & \textbf{79.3} \\
\noalign{\smallskip}
\hline
\noalign{\smallskip}
\end{tabular}
\caption{Number of K600 videos used and resulting accuracy for few-shot experiments with R3D-50 backbone.} 
\label{tab:fewshot}
\end{table}

\section{Metadata Analysis}

We present additional analyses and examples of the metadata in the WTS-70M dataset in Tables~\ref{tab:supp1}, \ref{tab:supp2}, and \ref{tab:supp3}, each of which shed some light on why titles are generally the most useful piece of metadata for sueprvision. Table~\ref{tab:supp1} shows the portion of unique instances for each metadata type. Titles have the highest proportion of unique instances, and are also the most useful signal for pre-training, implying that having many unique titles may be helpful. Table~\ref{tab:supp2} shows the distribution of lengths of each metadata type. Here, titles again show different characteristics from other forms of metadata; the median title contains 4 words, while the median description contains only 3 (despite descriptions being much longer on average). This indicates one other reason my titles might be the most useful, because longer metadata contains more information about the video. Table~\ref{tab:supp3} shows some examples of the most common titles and tags found in the dataset. We find that titles, even when repeated many times in the dataset, are generally informative about the content of the video. Other forms of metadata are less informative.

\setlength{\tabcolsep}{4pt}
\begin{table}[h]
\begin{center}
\begin{tabular}{lcc}
\toprule
Metadata & Num. Unique & \% Unique \\
\midrule [0.1em]
Titles & 43.0M & 61.5 \\
Descriptions & 29.3M & 41.9 \\
Tags & 34.0M & 48.6 \\
Channel Name & 21.0M & 29.9 \\
\bottomrule [0.1em]
\end{tabular}
\end{center}
\caption{Number of unique instances for each metadata type in WTS-70M. All metadata types contain repeats though some are repeated more often than others. Many channels are repeated, and we on average collect 3.3 videos per channel.}
\label{tab:supp1}
\end{table}

\setlength{\tabcolsep}{4pt}
\begin{table}[h]
\begin{center}
\begin{tabular}{lccccc}
\toprule
Metadata & Min & 25 & 50 & 75 & Max \\
\midrule [0.1em]
Titles & 0 & 2 & 4 & 6 & 158 \\
Descriptions & 0 & 0 & 3 & 12 & 4249 \\
Tags & 0 & 0 & 0 & 5 & 161 \\
Channel Name & 0 & 1 & 2 & 2 & 306 \\
\bottomrule [0.1em]
\end{tabular}
\end{center}
\caption{Quartiles of length (in words) of each metadata type. All have a long-tailed distribution, meaning that in extreme cases, the metadata may be hundreds or thousands of words long. However, all metadata types also contain examples which are empty or contain zero words. Titles are shortest in the most extreme cases, but longest in the median case.}
\label{tab:supp2}
\end{table}

\setlength{\tabcolsep}{4pt}
\begin{table}[h]
\begin{center}
\begin{tabular}{llc}
\toprule
Metadata & Text & Instances \\
\midrule [0.1em]
\multirow{10}{*}{Titles} & ``Free fire'' & 92K \\
 & ``Dance'' & 50K \\
 & ``Dancing'' & 47K \\
 & ``Baby'' & 34K \\
 & ``Bottle flip'' &  31K \\
 & ``Free Fire'' & 29K \\
 & ``Cute baby'' &  29K \\
 & ``Playing games'' &  27K \\
 & ``Games'' & 21K \\
 & ``Snow'' & 20K \\
\noalign{\smallskip}
\hline
\noalign{\smallskip}
\multirow{10}{*}{Tags} & ``PlayStation 4'' & 752K \\
 & ``Sony Interactive Entertainment'' & 695K \\
 & ``funny'' &  672K \\
 & ``video'' &  547K \\
 & ``mobile'' & 539K \\
 & ``YouTube Capture'' & 523K \\
 & ``\#PS4Live'' & 490K \\
 & ``how to'' &  467K \\
 & ``tutorial'' & 442K \\
 & ``fun'' & 371K \\
\bottomrule [0.1em]
\end{tabular}
\end{center}
\caption{Top ten most often-repeated titles and tags. For titles, these are descriptive and reflect the content of the video. For tags, these often contain automatically-generated metadata which reflect the method by which the video was uploaded.}
\vspace{10mm}
\label{tab:supp3}
\end{table}

\end{appendices}

\end{document}